%% file: main.tex
\documentclass{article}

\usepackage[utf8]{inputenc} 
\usepackage[T1]{fontenc}    
\usepackage{hyperref}       
\usepackage{url}            
\usepackage{booktabs}       
\usepackage{amsfonts}       
\usepackage{nicefrac}       
\usepackage{microtype}      
\usepackage{amsmath}
\usepackage{graphicx}
\usepackage{subcaption}
\usepackage[inline]{enumitem}
\usepackage[leftcaption]{sidecap}
\usepackage[dvipsnames]{xcolor}

\usepackage{color}
\usepackage{soul}

\usepackage[toc,page]{appendix}

\usepackage{hyperref}

\usepackage[accepted]{icml2021}

\icmltitlerunning{Understanding Invariance via Feedforward Inversion of Discriminatively Trained Classifiers}

\begin{document}

\twocolumn[
\icmltitle{Understanding Invariance via Feedforward Inversion\\of Discriminatively Trained Classifiers}



\icmlsetsymbol{equal}{*}

\begin{icmlauthorlist}
\icmlcorrespondingauthor{Piotr Teterwak }{piotrt@bu.edu}
\icmlauthor{Piotr Teterwak}{bu}
\icmlauthor{Chiyuan Zhang}{google}
\icmlauthor{Dilip Krishnan}{google}
\icmlauthor{Michael C. Mozer}{google,uc}
\end{icmlauthorlist}
\icmlaffiliation{google}{Google Research}
\icmlaffiliation{bu}{Presently at Boston University; work was begun while author was an AI Resident at Google Research}
\icmlaffiliation{uc}{University of Colorado, Boulder}

\icmlkeywords{Machine Learning, ICML}

\vskip 0.3in
]



\printAffiliationsAndNotice{}  

\begin{abstract}

A discriminatively trained neural net classifier can fit the training data perfectly if all information about its  input other than class membership has been discarded prior to the output layer. Surprisingly,  past research has discovered that some extraneous visual detail remains in the logit vector. This finding is based on inversion techniques that map deep embeddings back to images.  We explore this phenomenon further using a novel synthesis of methods, yielding a feedforward inversion model that produces remarkably high fidelity reconstructions, qualitatively superior to those of past efforts. When applied to an adversarially robust classifier model, the reconstructions contain sufficient local detail and  global structure that they might be confused with the original image in a quick glance, and the object  category can clearly be gleaned from the reconstruction. Our approach is based on BigGAN (Brock, 2019), with  conditioning on logits instead of one-hot class labels. We use our reconstruction model as a tool for exploring the nature of representations, including: the influence of model architecture and training objectives (specifically robust losses), the forms of invariance that networks achieve, representational differences between correctly and incorrectly classified images, and the effects of manipulating logits and images. We believe that our method can inspire future investigations into the nature of information flow in a neural net and can provide diagnostics for improving discriminative models. We provide pre-trained models and visualizations at \url{https://sites.google.com/view/understanding-invariance/home}.

\end{abstract}

\section{Introduction}

Discriminatively trained deep convolutional networks have been enormously 
successful at classifying natural images. During training, success is quantified
by selecting the correct class label with 
maximum
confidence. To the extent
that training succeeds by this criterion, the output must be invariant
to visual details of the class instance such as brightness changes, object pose, background configurations or small amounts of additive noise. Consequently, the net is encouraged
to discard information about the image other than the class label.
A dog is a dog whether the input image contains a closeup of a black puppy 
in a field or an elegant white poodle being walked on a city street. 

The successive layers of a convolutional
net detect increasingly abstract features with decreasing spatial specificity, 
from pixels to edges  to regions to local object components---like eyes and legs---to
objects---like dogs and cats \cite{zeiler2014visualizing}. %
Is is commonly believed that this sequence
of transformations filters out irrelevant visual detail in favor of 
information critical to discriminating among classes. This view is 
generally supported by methods that have been developed to invert 
internal representations and recover the visual information that
is retained by the representation in a given layer
\cite{mahendran2015understanding,dosovitskiy2016generating,dosovitskiy2016inverting,zhang2016augmenting,shocher2020semantic}.
Inversion of layers close to the input yield accurate 
reconstructions,  whereas inversion of deep layers typically 
results in a loss of visual detail and coherence. Attempts
to determine what visual information remains in the 
class output distribution, as expressed by the logits that are passed into 
the final softmax layer, have not been particularly compelling.
A notable attempt to invert the logits is the work of \citet{dosovitskiy2016generating}, which recovered colors, textures and the coarse arrangement of image elements from the logits. However, the reconstructions appear distorted and unnatural, and in the eleven examples shown in
\citet[][Figure 5]{dosovitskiy2016generating}, only about half of the object
classes are identifiable to a human observer. 
One would not confuse the reconstructions with natural images.
Although many visual details are present, the essence of the objects
in the original image is often absent.

In this paper, we demonstrate that 
the logit vector of a discriminatively trained network contains surprisingly rich information about not only the visual details of a specific input image, but also the objects and their composition in a scene. With a method that leverages a combination of previously proposed techniques  \cite{dosovitskiy2016generating,brock2018large}, 
we obtain remarkably high fidelity reconstructions of a source image from 
the logit vector of an ImageNet classifier. To give the reader a peek ahead, examine Figure~\ref{fig:reconstructions} and compare
the original image in column 1 of each set of five similar images to our reconstruction in column 2.
We show that the failure of previous efforts was not due to the loss of instance-specific visual information in the logits, but due to the less powerful inversion machinery. 
Apart from offering high quality reconstructions, our approach is computationally efficient and flexible for studying the reconstructions under various manipulations on the logits. We therefore leverage our method to explore the properties of logit representations across architectures and optimization methodologies; we particularly focus on comparing the properties of \emph{robust logits}, optimized with an adversarial training loop~\citep{goodfellow2014explaining}, and \emph{standard logits}, trained with a standard (non-robust) optimizer. Our contributions are as follows:

\begin{figure*}[p!]
  \centering
  \includegraphics[width=.975\linewidth]{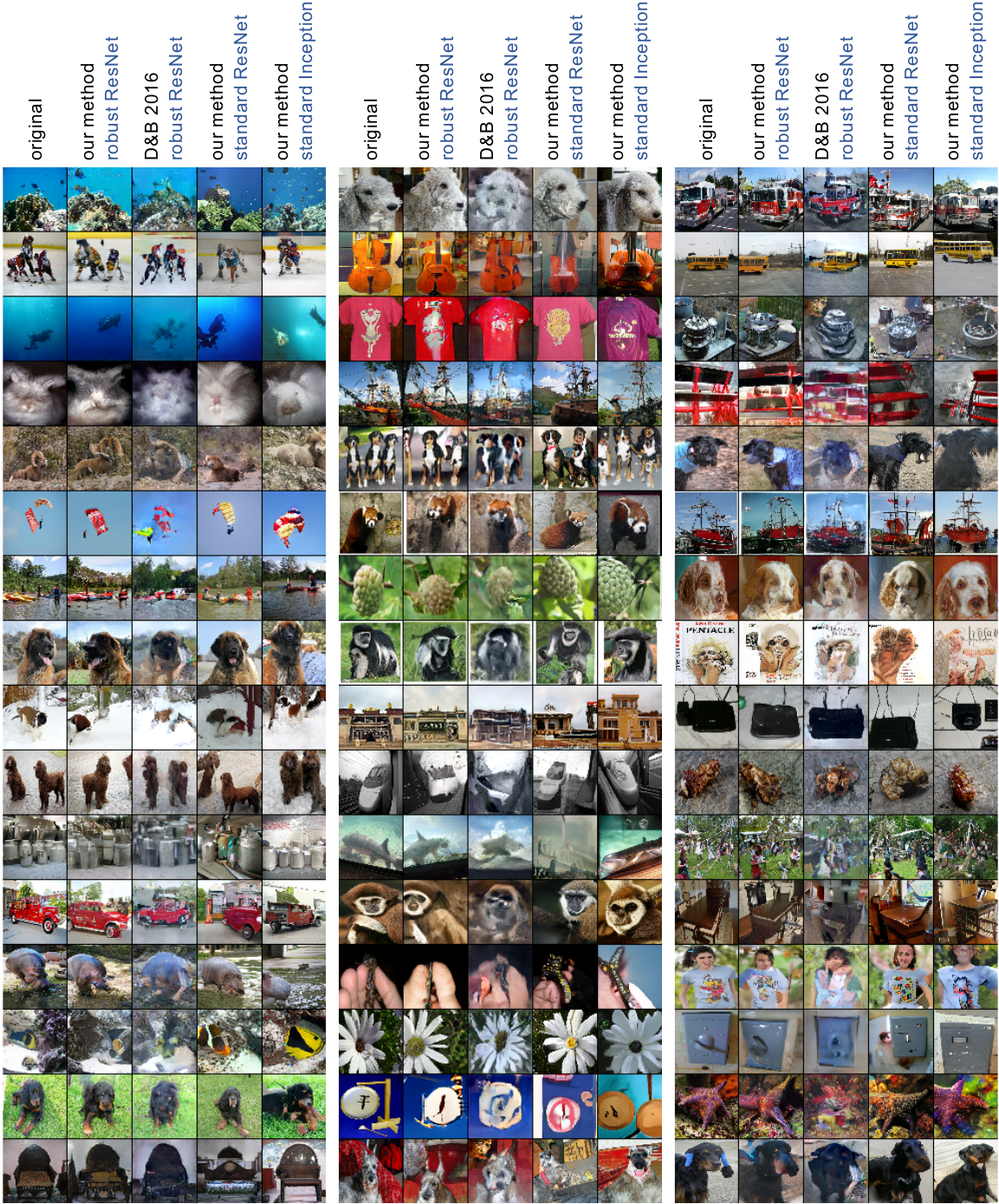}
  \caption{\small Original images and logit reconstructions.
  In each set of five images, the columns are:
  (1) the original image input to the classifiers,
  (2) reconstruction from our method using logits of a robust ResNet-152,
  (3) reconstruction from the method of \citet{dosovitskiy2016generating} using logits of a robust ResNet-152,
  (4) reconstruction from our method using logits of a standard (non-robust) ResNet-152, and
  (5) reconstruction from our method using logits of a standard (non-robust) Inception-V3.
  The images are selected at random from the test set with the only constraint that
  all classifiers produced the correct response to the images. These images were not used for training either the classifier or the reconstruction method.
  }
  \label{fig:reconstructions}
\end{figure*}

\begin{itemize}
    \item We improve on existing feature inversion techniques by leveraging conditional projection discriminators and conditional batch-norm. Compared to prior work \citet{dosovitskiy2016generating,dosovitskiy2016inverting}, this method generates higher qualitative reconstructions and is simpler to implement. 
    \item We show that both classifier architecture and optimization procedure impact the information preserved in logits of a discriminatively trained model. In particular, robust classifiers show significantly better reconstructions, suggesting that robust logits encode more object- and shape-relevant detail than non-robust logits. Further, a ResNet architecture
    appears to preserve more geometric detail than an Inception architecture does.
    \item We leverage our inversion technique to explore logit reconstructions for:
    \begin{enumerate*}[label=(\arabic*)]
        \item correctly classified images (and transforms that yield the same response)
        \item incorrectly classified images,
        \item adversarially attacked images,
        \item manipulations in logit space, including shifts, scales, perturbations, and interpolations, and
        \item out-of-distribution data.
    \end{enumerate*}
    \item Our experiments show that robust logits behave differently than non-robust logits. Most notably, our inversion model of robust logits, trained on ImageNet, can invert data from other datasets without retraining. This supports the view that adversarial training should be used in real-life scenarios when out of domain generalization is important.  
\end{itemize}

\section{Related research}

Methods developed to invert representations in classification networks 
fall into two categories: %
\emph{optimization based} and \emph{learning based}.
Optimization based methods perform gradient descent in the image space to determine 
images that yield internal representations similar to the representation 
being inverted, thus identifying an equivalence class of images insofar as 
the network is concerned. Back propagation through the classification
network is used to compute gradients in input space. 
For example, \citet{mahendran2015understanding} search over image space, $x \in
\mathbb{R}^{H \times W \times C}$, to minimize a loss of the form:
\begin{equation}
\mathcal{L}(x,x_0) = 
 || \Phi(x) - \Phi(x_0) ||_2 + \lambda \mathcal{R}(x) ,
\end{equation}
where $x_0$ is the original image, $\Phi(x)$ is the deep feature 
representation of input $x$,  $\mathcal{R}$ is a natural 
image prior, and $\lambda$ is a weighting coefficient. One drawback of this
method is that the solution obtained strongly depends on the random initialization of the optimization procedure.  With $\lambda=0$, the inverted representation does 
not resemble a natural image. 
\citet{engstrom2019learning} argued that training a model for adversarial robustness 
\citep{madry2017towards} provides a 
useful prior to learn meaningful high-level visual representations.
To make their argument, they reconstruct images from representation vectors from the penultimate layer of a robust model using the iterative
method. They showed reconstructions for a few examples and found that the recovered image
is less sensitive to the initial state of the gradient-based search, and that image-space gradients are perceptually meaningful.

Learning based methods use a separate training
set of \{logits, image pixels\} pairs to learn a decoder network that maps a logit vector to an image \cite{dosovitskiy2016generating,dosovitskiy2016inverting,nash2019inverting,rombach2020making}.
After training, image reconstruction is obtained via feedforward computation without expensive iterative optimization. 
For example, \citet{dosovitskiy2016inverting}
train a decoder network via a pixel reconstruction loss and an image prior.
To improve the blurry reconstructions, 
\citet{dosovitskiy2016generating} 
followed up with an approach most similar to ours, in which
they add an adversarial loss to the reconstruction and perceptual
losses, which substantially improves the reconstruction quality. We follow a similar approach, but make use of recent advances in generative modelling such as conditional projection discriminators and conditional batch normalization. These modification give higher quality results, and result in a model 
that is less sensitive to hyperparameter tuning.

\section{Models}
\begin{figure}[tb]
  \centering
  \includegraphics[width=\linewidth]{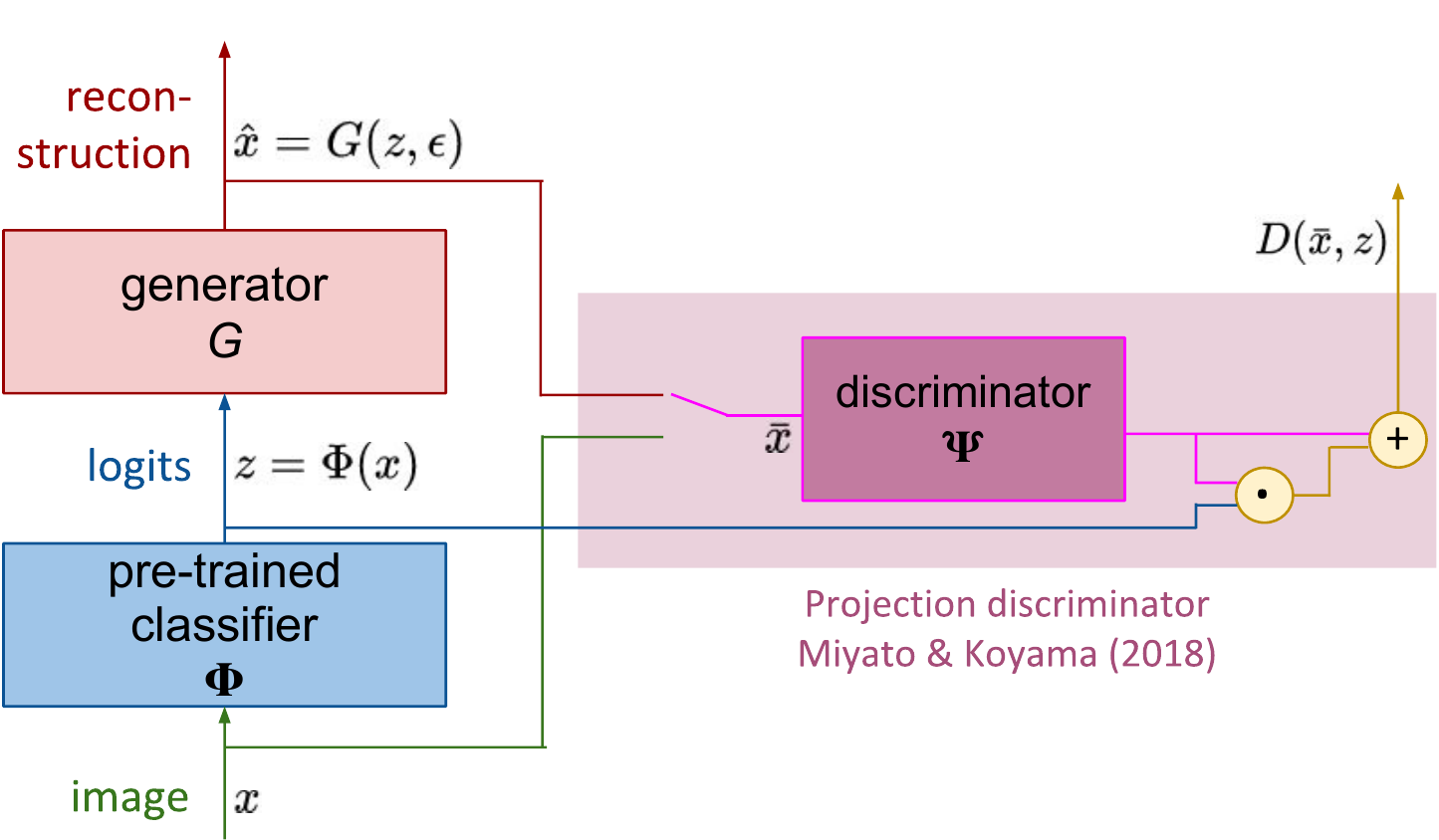}
  \caption{A pretrained and fixed-weight classifier provides logit vectors which are used
  by a conditional GAN \cite{MirzaOsindero2014} to generate reconstructions. The discriminator is based on the projection discriminator of \cite{miyato2018cgans}, which is fooled when both
  the reconstructed image looks natural and is a match to the logit vector of the real image, $x$.}
  \label{fig:cGANfigure}
\end{figure}

We adopt a learning based approach for reconstructing from the logits. Specifically, we train conditional GANs \cite{MirzaOsindero2014} to synthesize images. Given an original image $x$ that is passed through a pre-trained classifier $\Phi$ to obtain an $n_c$-dimensional logit vector, 
$z=\Phi(x)$,
our method focuses on obtaining a reconstruction of the input, $\hat{x}$, from a generative model $G$ that is conditioned on $z$:
$\hat{x}=G(z,\epsilon)$,
where $\epsilon \sim \mathcal{N}(0,1)$ is a 120-dimensional noise vector that is used to obtain diversity in the generator output. 

We build on the work of \citet{dosovitskiy2016generating} by using state-of-the-art classifiers and a powerful, elegant adversarial model. Specifically, we leverage batch-norm generator conditioning---first used for style transfer 
\cite{dumoulin2016learned,de2017modulating}, and later used in GANs 
\cite{miyato2018cgans,brock2018large}; and projection discriminators---first 
introduced in \citet{miyato2018cgans} and further popularized by BigGAN \cite{brock2018large}. Instead of conditioning the model components using a 
one-hot class representation, we condition on the target logit distribution, 
$\Phi(x_0)$. Such feature-based conditioning of the discriminator is similar to 
Boundless \cite{Teterwak_2019_ICCV}, and the feature conditioning of the generator 
is similar to SPADE \cite{park2019SPADE}, also known as GauGan. 

The generator network is trained to synthesize images that can fool a discriminator network. The weights of both networks are optimized jointly. The discriminator, $D(\bar{x},z)$, takes either a real image,
$\bar{x} \equiv x$, or its corresponding generated image, 
$\bar{x} \equiv \hat{x}$, along with the logit vector $z$ which 
is either produced by the classifier for the real image, 
or is used  to synthesize the generated image.
The discriminator 
outputs a scalar, a large positive value when 
$\bar{x}$ is real and a large negative value when $\bar{x}$ is 
synthetic.

The discriminator consists of two terms, one 
of which makes the judgment based on whether the image is 
naturalistic and the other based on whether the image would have produced 
the given logits.  Inspired by the projection discriminator of 
\citet{miyato2018cgans}, we use
\begin{equation}
    D(\bar{x},z) = \left( w_1 + W_2z \right)^\top \Psi(\bar{x}),
\end{equation}
where 
$\Psi(\cdot)$ is a deep net
that maps to a $n_d$-dimensional feature vector; and
$w_1 \in \mathbb{R}^{n_d}$, 
$W_2 \in \mathbb{R}^{n_d \times n_c}$. 
The $w_1$ term helps discriminate
images based on whether they appear real or synthetic, and the $W_2$ term
helps to discriminate images based on whether or not they are consistent
with the logit vector $z$.
The overall architecture is shown in Figure~\ref{fig:cGANfigure}.

For an image $x$ and its corresponding logit vector $z = \Phi(x)$, 
we have adversarial losses for the discriminator $D$, and the generator $G$:
\begin{equation*}
\begin{aligned}
	\mathcal{L}_D \hspace{-.01in} &= \mathbb{E}_{x,\epsilon} \hspace{-.05in} \left[
	\max \left(-1, D\left(\bar{G}(z, \epsilon), z\right)\right) 
	-\min \left(1, D(x,z)\right)  \right],\\
	\mathcal{L}_G &= -\mathbb{E}_{x,\epsilon} \left[ \bar{D}(G(z, \epsilon), z) \right],
\end{aligned}
\end{equation*}
with $z = \Phi(x)$, generator noise distribution $\epsilon 
\sim \mathcal{N}(0,1)$, and $\bar{G}$ and $\bar{D}$ denoting
the parameter-frozen generator and discriminator, respectively.

The discriminator is optimized to distinguish real and synthetic images
based both on the images themselves and the logit vector. 
As a result, the discriminator is driven to distill the
classification network and then apply a form of adversarial perceptual
loss \cite{Johnson2016Perceptual}.
The generator is optimized to fool the discriminator by synthesizing images that 
are naturalistic and consistent with the given logit vector.
Consequently, the discriminator must implicitly learn the mapping performed by
the classifier $\Phi(x)$.

\citet{dosovitskiy2016generating} were able to approximately invert representations by using a loss with three terms---adversarial, perceptual, and pixel reconstruction---which requires hyperparameter tuning. The approach we present has the advantage of using a single loss term and thereby avoiding tuning of relative loss weights.

Most of the previous conditional GANs uses the one-hot class label vector for conditioning the generators~\citep{brock2018large}. Our generator $G$ conditions on the logic vector $z$ instead. 
Following \citet{brock2018large}, we do not treat the conditioning vector $z$ as a conventional input layer for $G$, 
but rather use $z$ to modulate the operation of each batch-norm layer for a given channel $k$ and a given layer $l$
in $G$ as follows:
\begin{equation}
y'_{lk} = \frac{y_{lk} - \mathbb{E}[y_{lk}]}{\mathbb{S}[y_{lk}]} \gamma_{lk}(z)  +
\beta_{lk}(z)
\end{equation}
where $y$ and $y'$ are the layer input and output; 
$\gamma(.)$ and $\beta(.)$ are differentiable functions of $z$, which we implement as two-layer MLPs;
and the expectation $\mathbb{E}$ and standard deviation $\mathbb{S}$ are computed over all units in channel $k$
over all inputs in the batch of examples being processed.

\section{Methods}
We explore three pretrained ImageNet models:
\href{https://github.com/tensorflow/models/tree/master/research/slim}{ResNet-152-V2}
\citep{he2016deep},
\href{https://tfhub.dev/deepmind/local-linearity/imagenet/1}{Robust ResNet-152-V2}
\citep{qin2019adversarial}, and
\href{https://tfhub.dev/google/imagenet/inception_v3/classification/4}{Inception-V3}
\citep{szegedy2016rethinking}.
The implementations we used are linked under the model names.

We train generators to produce $64\times64$ images, and therefore use $64\times64$ images for input. We modify the BigGAN implementation in the 
\href{https://github.com/google/compare_gan}{\texttt{compare\_gan} code base}
to support our new conditioning, with the configuration as described in the Supplementary Materials. 
We train the generator on TPUv3 accelerators using the same ImageNet~\citep{russakovsky2015imagenet} training set that the classifiers are trained on, and evaluate using the test set. 

We also replicated the method of \citet{dosovitskiy2016generating}, upgrading the AlexNet classifier they were inverting to a 
state-of-the-art model.
We were able to retrain their method, using
their official Caffe \cite{jia2014caffe} implementation, to invert the logits from the robust ResNet-152.

\section{Results}

\subsection{Comparing reconstruction methods}
We begin by comparing reconstructions from our method with those from the method of \citet{dosovitskiy2016generating}.
These reconstructions are shown in columns 2 and 3 of the image sets in Figure~\ref{fig:reconstructions}, respectively, and can be compared to the 
original images in column 1. Both methods are trained using robust ResNet-152 logits. The images shown in the Figure were not used
for training, either the classifier or the reconstruction method.  Although both methods capture color and texture
well, and both methods seem to generate sensible reconstructions in a quick glance for scenes that are basically textures---such as
the underwater seascapes---the \citeauthor{dosovitskiy2016generating} method fails in recovering object shape, details, and the relationships
among parts. For example, distinguishing one dog species from another is quite challenging, and the rabbit is a fuzzy blur.

\subsection{Reconstruction from different classifiers}

We now turn to exploring reconstructions using our method, trained for different classifiers. In particular we direct the
reader to columns 2, 4, and 5 of each image set, which correspond to reconstructions using 
a robustly trained ResNet-152 \citep{qin2019adversarial}, 
a standard ResNet-152 \cite{he2016deep},  and
a standard Inception-V3 \cite{szegedy2016rethinking}. 
Between columns 2 and 4 we can compare  adversarial robust training to standard training with the same architecture,
and between columns 4 and 5 we can compare different architectures under the same (standard) training procedure.

Before examining these detailed differences, we note that overall the reconstructions
from all models capture significant detail about the visual appearance of the image, more than just
its class label. In a quick glance, the reconstruction would convey very similar information as the original image.
In general, color, texture details, and backgrounds are preserved. However, other information is not preserved, 
including: left-right (mirror) orientation, precise positions of the image elements, details of the background
image context, and the exact number of instances of a class (e.g., the hockey players or the black, 
white, and brown dogs). The loss of left-right reflection and image-element positions may be due to the fact
that these classifier models are trained with the corresponding data augmentations.
From those reconstructions we conclude that the logit vector of a discriminatively trained classifier network indeed contains rich information about the input images that could be extracted to reconstruct the input with high fidelity in visual details.

\citet{engstrom2019learning} showed that optimization-based feature inversion is 
significantly improved when using robustly trained classifier models. 
The reason could be
either that the input-space gradients are more perceptually relevant, or that the robust features actually encode more information, or both. Being a learning-based method, our model is not dependent on the input-space gradients. As a result, when we invert robust logits,  we can answer the question whether adversarial optimization encodes more instance level information than standard optimization. Examining Figure~\ref{fig:reconstructions}, the robust model reconstructions (column 2 of the image sets)
match the original image (column 1) better than the corresponding non-robust model reconstructions (column 4), capturing both local and 
global detail of the images. For example, the fourth-from-bottom row of the middle collection of images shows a small animal 
on human skin more clearly;  and the cello in the second row is more discernible in the robust model than in the non-robust model. Therefore our results are fully in agreement with prior work such as \citet{engstrom2019learning} and \citet{santurkar2019image}: robust models are indeed superior to non-robust models in terms of information captured in the logit layer. 

Comparing the two non-robust models, ResNet-152 (column 4) and Inception V3 (column 5), the reconstructions from ResNet seem to be truer to the
original image, but the reconstructions from Inception are often closer to photorealism.  For example, in the middle collection of images, the red 
t-shirt is reconstructed as purple with a very different design. And in the first row, middle collection, the ResNet dog better matches the original 
than the Inception dog. The Inception images are more stereotypical. Interestingly, ResNet-152 achieve much better classification performance than Inception. It is a bit surprising that a model that better at classification actually retains richer information relevant to per-instance visual details in the logit vectors.

These findings are further supported by a  small-scale human evaluation (4 subjects). Each subject was asked to perform three two-alternative forced choice tasks, each task with 64 images. In each task, the subjects compared pairs of reconstructed images to a ground truth image and were asked to indicate which of the reconstructions was closer to ground truth.
As the results in Table \ref{tab_2afc} indicate: (1) reconstructions by our method is overwhelmingly preferred to that of \citet{dosovitskiy2016generating} for the same architecture; (2) a robust ResNet is overwhelmingly preferred to a non-robust ResNet with the same loss; and (3) among non-robust networks, ResNet is slightly preferred over Inception. 

For each of the four networks, we also computed the LPIPS metric \cite{zhang2018unreasonable} over a set of images (Table \ref{tab_lpips}). We see that the metric generally supports the preferences of Table \ref{tab_2afc} except for the comparison between our method and that of 
\citet{dosovitskiy2016generating}, which slightly favors the latter. We believe the reason for this discrepancy is that \citet{dosovitskiy2016generating} train by minimizing a perceptual loss very similar to that of LPIPS. This highlights the bigger challenge of creating metrics that are not also used as optimization criteria. 

\begin{table}
{\small
\centering
\begin{tabular}{ | c |}
\hline
 Our Robust ResNet vs. D\&B Robust ResNet \\ {\bf Ours: 87.5\%}; D\&B: 12.5\% \\ 
 \hline
 Our Robust ResNet vs. Our Non-Robust ResNet \\ {\bf Robust: 76.5\%}; Non-robust:  23.5\% \\  
 \hline
 Our ResNet vs. Our Inception \\ {\bf ResNet: 52.7\%}; Inception: 47.3\% \\ 
\hline
\end{tabular}
\caption{Two-Alternative Forced Choice (4  subjects, 64 images): ``Which reconstruction is closer to the ground truth?"}
\label{tab_2afc}
}
\end{table}

\begin{table}
{\small
\centering
\begin{tabular}{ | c | c |}
\hline
 Robust Ours & 0.3138 \\ 
 \hline
 Robust D+B & 0.3092  \\  
 \hline
 Non-robust resnet & 0.3755 \\ 
 \hline
Inception-V3 & 0.3881   \\
\hline
\end{tabular}
\caption{LPIPS Between Input and Reconstruction (lower is better)}
\label{tab_lpips}
}
\end{table}

\subsection{Visualizing Model Invariances}

\begin{figure}[tb!]
  \centering
  \begin{subfigure}[b]{0.47\linewidth}
    \includegraphics[width=\linewidth]{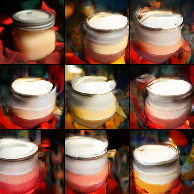}
  \end{subfigure}
  \hspace{.1in}
  \begin{subfigure}[b]{0.47\linewidth}
    \includegraphics[width=\linewidth]{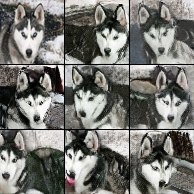}
  \end{subfigure}
  \caption{Variation in reconstructions due to noise resampling for Robust ResNet-152. The upper-left tile is the input.}
  \label{fig:resample}
\end{figure}

\begin{figure}[tb!]
  \centering
  \begin{subfigure}[b]{0.47\linewidth}
    \includegraphics[width=\linewidth]{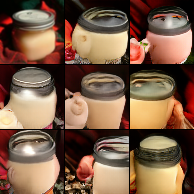}
  \end{subfigure}
  \hspace{.1in}
  \begin{subfigure}[b]{0.47\linewidth}
    \includegraphics[width=\linewidth]{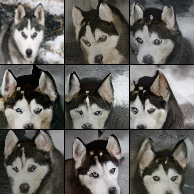}
  \end{subfigure}
  \caption{Variation in reconstructions due to noise resampling for non-robust ResNet-152. The upper-left tile is the input.}
  \label{fig:resample_non_robust}
\end{figure}

\begin{figure}[tb!]
  \centering
  \begin{subfigure}[b]{0.47\linewidth}
    \includegraphics[width=\linewidth]{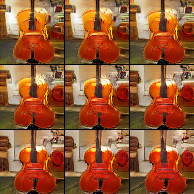}
  \end{subfigure}
  \hspace{.1in}
  \begin{subfigure}[b]{0.47\linewidth}
    \includegraphics[width=\linewidth]{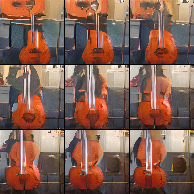}
  \end{subfigure}
  \caption{Noise interpolation for robust (left) and non-robust (right) ResNet-152. We linearly interpolate between random noise vectors. The top-left image corresponds to one noise sample, the lower-right corresponds to another, and all others are linear interpolates arranged left-to-right, top-to-bottom. We find that the reconstruction varies significantly less across noise inputs for the robust model. }
  \label{fig:select_noise_interpolate}
\end{figure}

Next, we explore the effects of resampling generator noise. 
The generator takes as input a Gaussian noise vector, in addition to the logit vector from the pretrained classifier.
The noise primarily captures non-semantic properties (Figure \ref{fig:resample}, \ref{fig:resample_non_robust}),
mostly small changes in shape, pose, size, and position. These properties reflect information that the classifier
has discarded from the logit vector because the generative model considers all of them to be sensible reconstructions of the same
logit vector. One particularly interesting invariance is left-right mirroring; our model frequently generates horizontal flips of an 
image, but not vertical flips. For example, the dog's body is sometimes on the left and sometimes on the right of the face. We do note, however, that the noise resampling has a greater effect on the non-robust model than on the robust model. This is easily seen in Figure \ref{fig:select_noise_interpolate}, where we linearly interpolate between two noise samples. The robust model is much more stable along the axis of interpolation. We provide many more examples in the Supplementary Materials. 

It is interesting to observe invariances being captured in the logits and the generator recovering them via resampled input noises. %
However, more subtle invariances may not be captured if the discriminator
has not learned to look for certain forms of variation as a cue.

\subsection{Reconstructing incorrectly classified images}

\begin{figure}[b!]
  \centering
  \includegraphics[width=\linewidth]{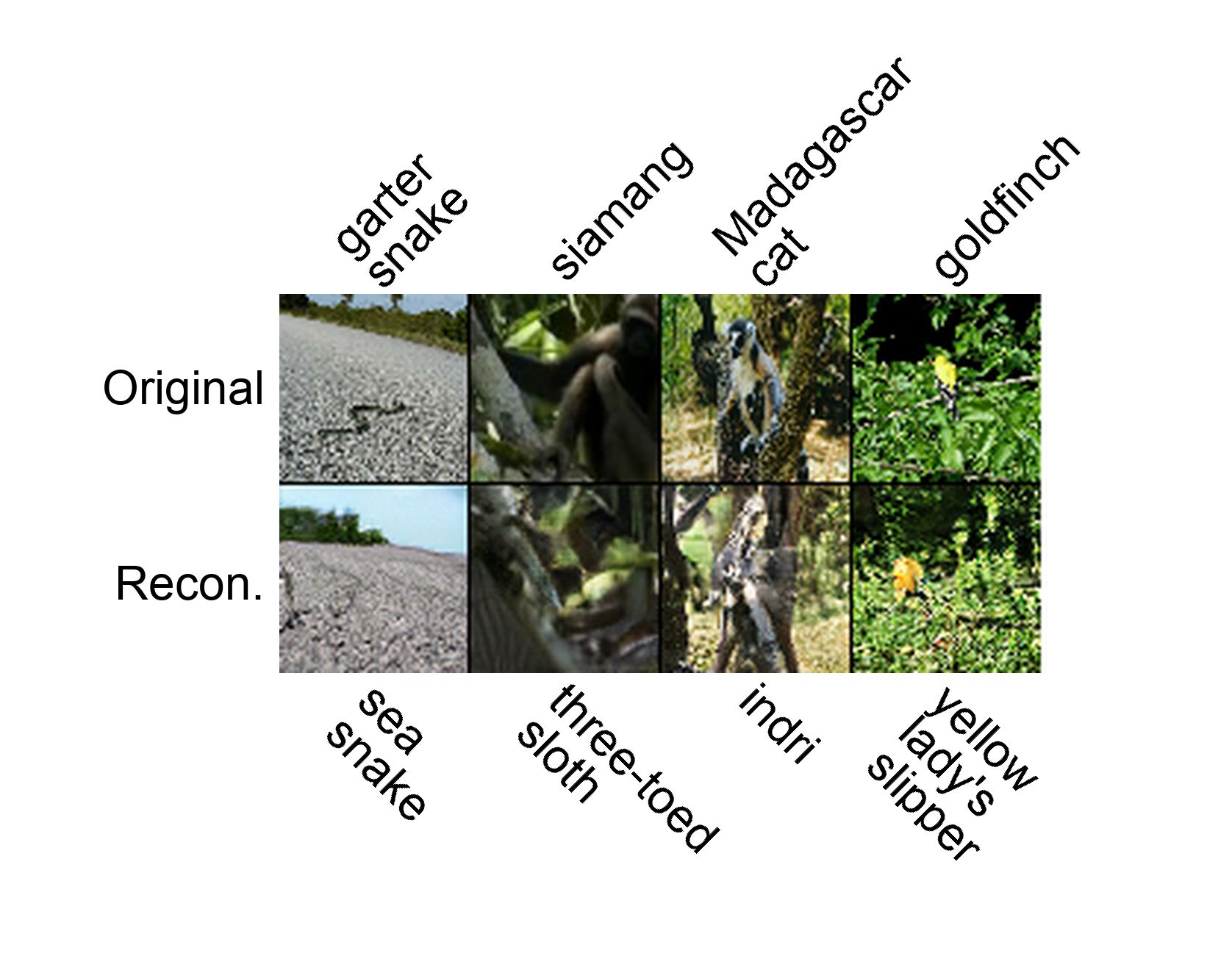}
  \caption{ Reconstruction of  incorrectly classified samples using robust ResNet-152. Surprisingly, even incorrectly classified samples are reconstructed faithfully.}
  \label{fig:correct_incorrect}
\end{figure}

The samples we show in Figure~\ref{fig:reconstructions} are of correctly classified images. Does the correctness of classification have
a significant impact on the nature of reconstruction?
If a sample is misclassified because the net `sees' a sample as belonging to a different class,
the reconstructions may reveal a categorical shift to the incorrect class. One might therefore expect
reconstructions of correctly classified samples to be more veridical than those of incorrectly classified samples.
Figure \ref{fig:correct_incorrect} shows that even incorrectly classified samples from the ImageNet test set are faithfully 
reconstructed. With this, we infer that incorrect classifications are due to the network drawing flawed decision boundaries
rather than a semantically flawed embedding. We provide more samples for both the robust and non-robust models in the Supplementary Materials. 

\begin{figure}[tb!]
  \centering
  \begin{subfigure}[b]{.5249\linewidth}
    \includegraphics[width=\linewidth]{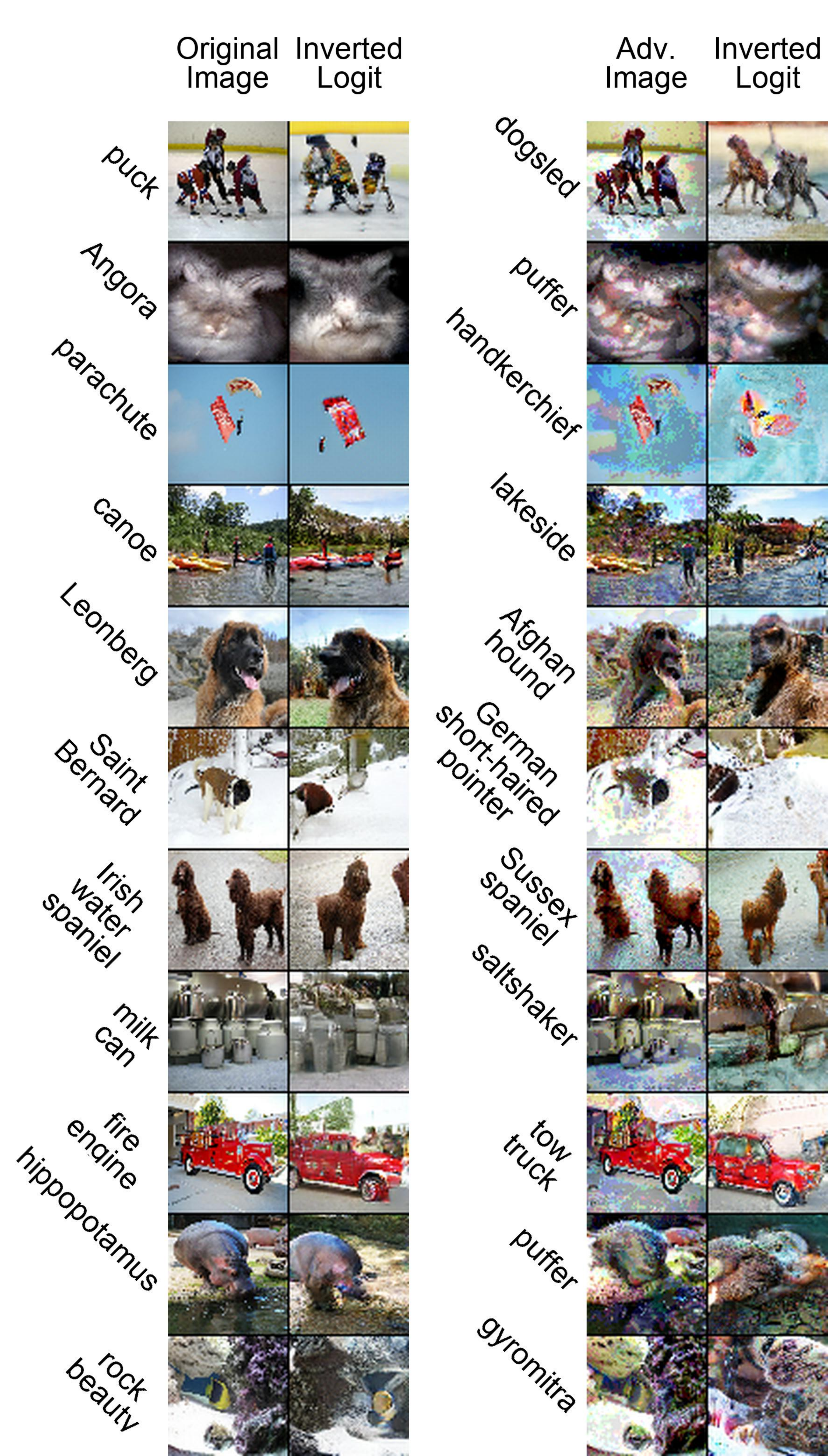}
   \caption{Robust}
  \end{subfigure}
  \hspace{.12in}
  \begin{subfigure}[b]{.41835\linewidth}
    \includegraphics[width=\linewidth]{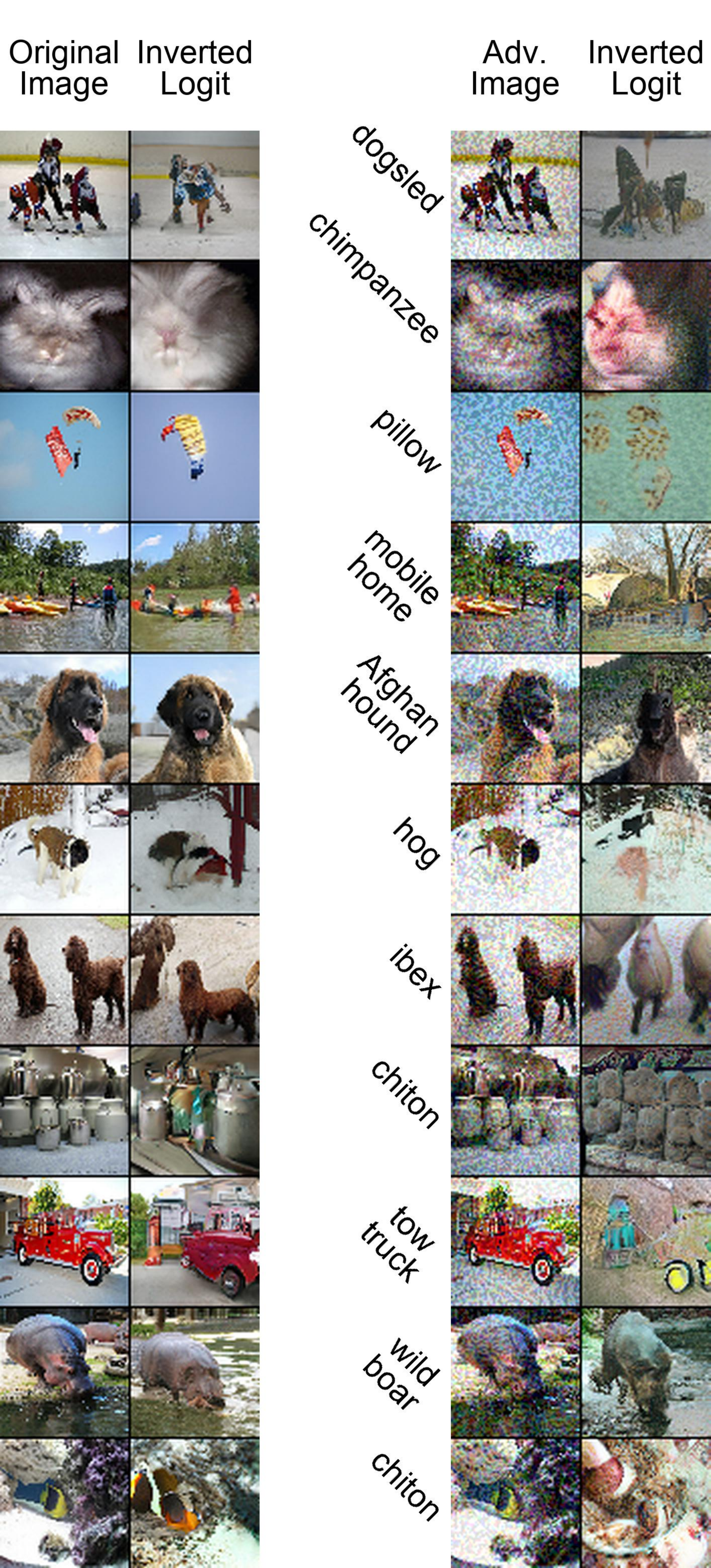}
    \caption{Non-robust}
  \end{subfigure}
  \caption{Reconstructions of adversarially attacked images using the FGSM method \cite{goodfellow2014explaining}. 
  Each column of image pairs consists of the input and reconstructed image. The left two columns are from robust ResNet and the right 
  two columns non-robust ResNet. Each pair of columns consists of the original and adversarial images and reconstructions. The true labels are on the very left, and the predicted labels after attack are also shown.}%
  \label{fig:adversarial}
\end{figure}

We turn now to another kind of incorrectly classified sample: adversarial examples \cite{goodfellow2014explaining}, small perturbed version
of correctly classified images that result in incorrect classification. We use FGSM~\citep{goodfellow2014explaining} to generate adversarial examples, with attack strength
$\epsilon = 0.1$, meaning that no pixel deviates by more than $\epsilon$ from the source image. 
Figure~\ref{fig:adversarial} shows the original and adversarial images alongside their reconstructions for the robust (left two columnns)
and non-robust (right two columns) ResNet models. The correct and incorrect labels are shown beside the images.
We selected images for which successful attacks could be found for both robust and non-robust models.
Unsurprisingly, the robust model is less sensitive to attacks. 
Whereas reconstructions of adversarial images in the robust model appear not to lose significant fidelity relative to reconstructions
of the original images, the same is not entirely true for the non-robust model. Take the fire
engine (3rd row from the bottom) as an example. The adversarial attack leads to a reconstruction in which the vehicle changes color and shape and no longer
looks like a fire engine. For the robust model, the adversarial images seem to be reconstructed about as well as ordinary incorrectly
classified images (Figure~\ref{fig:correct_incorrect}); in both cases, visual elements and shapes remain largely intact despite 
incorrect classification.

\subsection{Logit manipulations}

In this section, we explore how three manipulations of the logits affect reconstructions:  
\emph{logit shifting}, in which a constant is added to  each element of the logit vector; 
\emph{logit scaling}, in which each element of the logit vector is multiplied by a constant;  and
\emph{logit perturbation} in which i.i.d. Gaussian noises are added to each element of the logit vector. We hold the noise input of the generator constant when manipulating the logit for each sample. 

\begin{figure}[bt!]
  \centering
  \includegraphics[width=\linewidth]{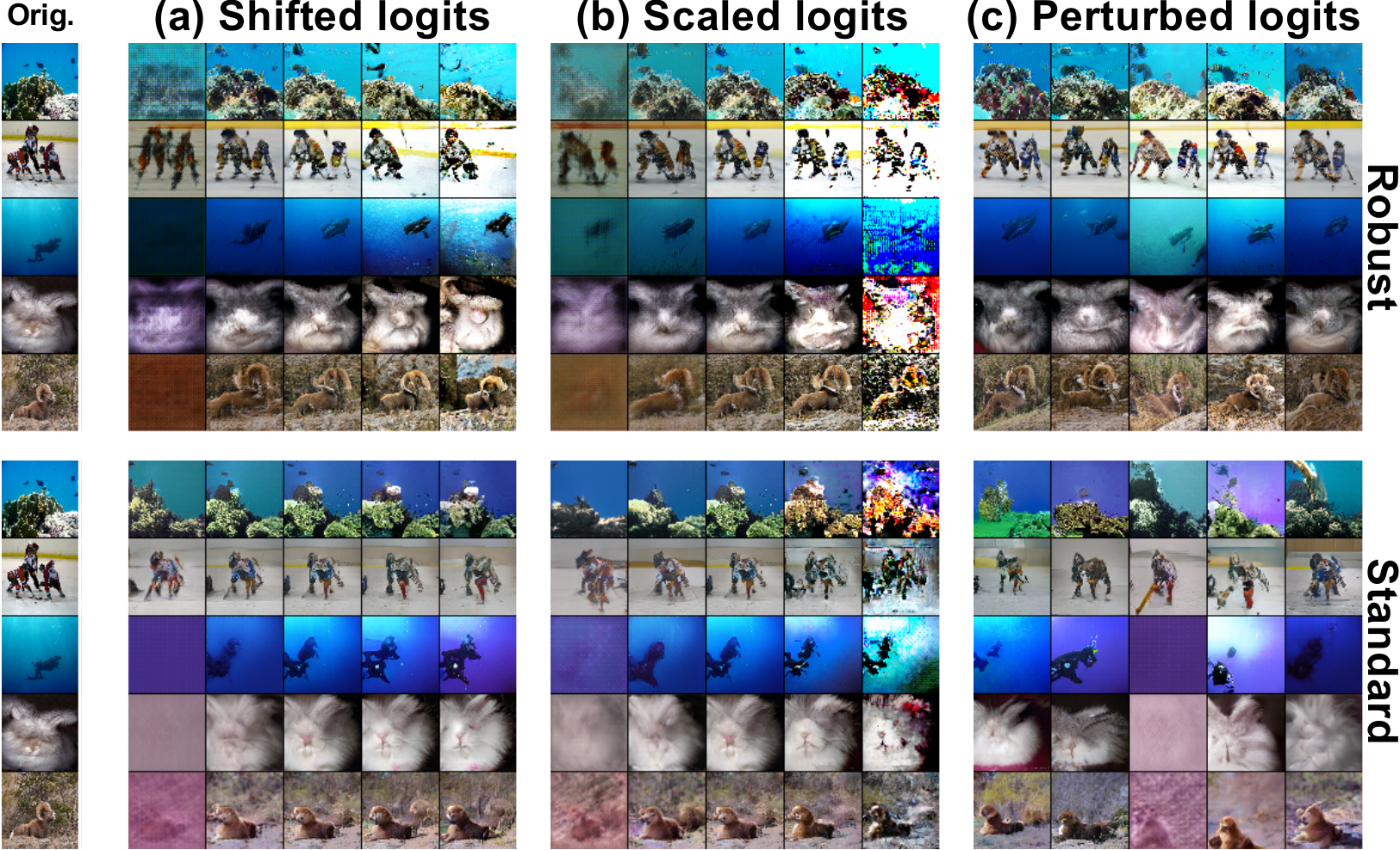}
  \caption{Reconstructions formed by shifting, scaling, and perturbing logits. The upper set of images is from robust ResNet-152 logits; the lower set is from standard ResNet-152 logits.}
  \label{fig:shift_and_scale}
\end{figure}

Figure~\ref{fig:shift_and_scale}a illustrates logit shifting. The five columns are reconstructions of an original image (left side of Figure) with each logit in the vector shifted by a constant. 
The upper and lower sets of images correspond to robust and standard (non-robust) ResNet-152 models.
For robust ResNet-152, the constants for the five columns are $-0.30$, $-0.15$, $0.0$, $0.15$, and $0.30$; 
for standard ResNet-152, the constants are $-0.1$, $-0.05$, $0.0$, $0.05$, and $0.1$. For larger shifts, reconstructions 
from standard ResNet lose meaningful contents.

For the robust model, the manipulation primarily affects contrast and sharpness of the reconstructed image but also has subtle effects on shape.
For example, in the hockey scene, the three players appear to morph into one with larger shifts. 
The effect is much less pronounced in the non-robust model, where there are some color changes with positive shifts and content suppression for large negative shifts.

Because a softmax classifier's output is invariant to logit shifting (the shifts are normalized out), there is no training 
pressure for the classifier to show any systematic relationship between image features and logit shifts.
It must thus be an inductive bias of the training procedure that image contrast and sharpness is reflected on offsets to the logits \cite{Scott2021}.
We refer the reader to the Supplementary Materials for additional experimental support for the hypothesis suggested by the reconstructions. We show that directly manipulating brightness of an image results in shifted robust logits.

Figure~\ref{fig:shift_and_scale}b illustrates logit scaling. The five columns are reconstructions with each logit in the vector scaled by a constant,
where the constants for the five columns are $10^{-0.3}$, $10^{-0.15}$, $10^0$, $10^{0.15}$, and $10^{0.3}$. Surprisingly, for the robust model this manipulation also
affects reconstruction contrast and sharpness, almost exclusively. Scaling affects the output confidence distribution: the larger the scaling factor,
the more binary model outputs become. Sensibly, the robust classifier has lower confidence for blurry low contrast images. The scaling manipulation appears
to affect \emph{only} contrast and sharpness. During training, the classifier loss \emph{is} sensitive to changes in the logit scale. Consequently it's 
somewhat surprising that scale encodes only contrast and sharpness, and content is well preserved across scale changes.
The robust classifier appears to use the embedding \emph{direction} to represent image content.

In the non-robust model, apart from the extreme ends of the scaling, there is no significant change in brightness. Furthermore, unlike in the robust model, there do seem to be slight changes in content. For example, the coral reef in the top row changes shape. Therefore, for non-robust classifiers, embedding direction \textit{and} scale encode content.

In Figure~\ref{fig:shift_and_scale}c, we show reconstructions in which the the logits are perturbed by i.i.d. Gaussian noise $\mathcal{N}(\mu=0,\sigma^2=0.55)$.
For both robust and non-robust models, image content is affected by noise. For the robust model, the content is not so much affected that one could not group together the reconstructions from the same
underlying logit vector. However, for the non-robust model the content changes are much larger; indicating that non-robust logits are much more tightly grouped in the output space.  We verify this quantitatively by selecting 10 random classes and computing per-class means of validation sample logit vectors. We then measure $l_2$ distances between samples from each class to their class mean. For the non-robust model, the mean $l_2$ distance is $42.33$, compared to $43.78$ for the robust model, supporting the hypothesis that non-robust logits are grouped more tightly. We note that noise added to the logits has a different effect than providing noise sample $\epsilon$ to the generator. Additive logit noise changes content, whereas $\epsilon$ primarily affects pose and location.

\begin{figure}[bt]
  \centering
  \begin{subfigure}[b]{0.49\linewidth}
    \includegraphics[width=\linewidth]{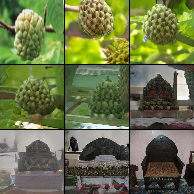}
  \end{subfigure}
  \begin{subfigure}[b]{0.49\linewidth}
    \includegraphics[width=\linewidth]{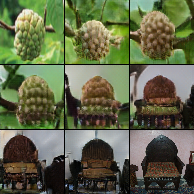}
  \end{subfigure}
  \caption{We compare non-robust logit interpolation (left) and robust logit interpolation (right). The upper left and lower right images in each block are from the ImageNet test set. The intermediate images are obtained by reconstructing interpolations between the logit vectors of these two images.}
  \label{fig:interpolate}
\end{figure}

Given the similarity of reconstructions from nearby points in logit space, we also explored the effect of interpolating between the logit vectors of two
distinct images. Figure~\ref{fig:interpolate} shows two sample reconstruction sequences, formed by linear interpolation between logit vectors, one from the robust model and one from the non-robust model. Although we observe a smooth semantic continuum in which natural looking images are obtained at each interpolation point for both, the robust model has a smoother sequence. 
We show many more examples in the supplementary materials. 
Other researchers studying image reconstruction from logits have also observed smooth interpolations
\citep[e.g.,][]{dosovitskiy2016generating},
though we appear to be the first to show more organization in the space for
robust models.

\subsection{Reconstructing out-of-distribution data}

\begin{figure}[bt!]
  \centering
  \begin{subfigure}[b]{0.325\linewidth}
    \includegraphics[width=\linewidth]{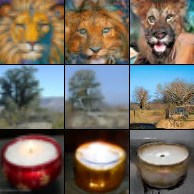}
  \end{subfigure}
  \begin{subfigure}[b]{0.325\linewidth}
    \includegraphics[width=\linewidth]{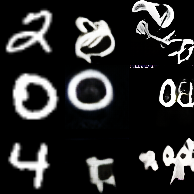}
  \end{subfigure}
  \begin{subfigure}[b]{0.325\linewidth}
    \includegraphics[width=\linewidth]{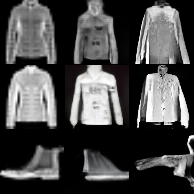}
  \end{subfigure}
  \caption{Reconstruction of CIFAR-100, MNIST, and FashionMNIST images with a model trained on ImageNet. The left column is input, middle is Robust ResNet-152 reconstruction, right is standard ResNet-152 reconstruction.} %
  \label{fig:mnist_cifar}
\end{figure}

To further understand how classifiers behave on out-of-distribution (OOD) data, we reconstruct samples from non-Imagenet datasets using a model  trained on ImageNet. Both the classifier used to construct the logits \textit{and} the inversion model are trained on ImageNet. ImageNet pre-training for transfer learning is an extremely common and successful method, so it's natural to ask \textit{what} about the target dataset is embedded in the features. In Figure \ref{fig:mnist_cifar}, we compare the robustly trained ResNet-152 with the standard one on CIFAR-100~\citep{Krizhevsky09learningmultiple}, MNIST~\citep{lecun2010mnist}, and FashionMNIST~\citep{DBLP:journals/corr/abs-1708-07747}. It is notable that the robust model offers substantially better reconstructions than the non-robust model. 
This ability to encode OOD data strongly supports claims of \citet{salman2020adversarially} that robust features are better for transfer learning.

\section{Discussion}

We summarize our results as follows.
\begin{itemize}
    \item We obtain remarkably high fidelity reconstructions, which allow the object class to be determined as well as visual
    detail orthogonal to the class label. In contrast to current state-of-the-art reconstruction methods 
    \cite{dosovitskiy2016generating,engstrom2019learning}, our model preserves global coherence as well as local features.
    \item Subjectively, robust ResNet produces better reconstructions than non-robust ResNet, suggesting that the
    adversarial training procedure is effective in preserving features that human observers identify as salient in
    an image, even if non-class-related.
    \item Architecture matters. ResNet seems to better preserve visual information than Inception. It seems likely that the short
    circuit linear connections of ResNet allow low level information to be propagated forward.
    \item We do not see a qualitative difference in reconstruction fidelity for incorrectly classified images. 
    \item For a robust ResNet-152, both logit shifts and rescaling have have a similar effect---they influence the contrast, sharpness, and brighness of reconstructions.
    The relationship for non-robust ResNet is similar but weaker.
    \item The correspondence between logit space and image space is smooth, such that small perturbations to the logits 
    yield small perturbations to reconstructions, and interpolating between logits produces reasonable image interpolations. The interpolation produces a smoother sequence for the robust model.
    \item The robust ResNet-152 encodes OOD data such as CIFAR and MNIST much more faithfully than non-robust ResNet-152, giving a clue as to why robust models are better for transfer learning. 
\end{itemize}

The degree to which the logit vector is invertible seems quite surprising. After
all,  perfectly discriminative networks should retain only class-relevant information
and should be invariant to differences among instances of a class.

Future work should focus on how to leverage what we learn to design better systems. If robust classifiers result in more invertible features, is it also true that invertible features result in more robust classifiers? Can we use decoded interpolated logits as a form of semantic MixUp \cite{zhang2017mixup}? Additionally, it would be interesting to inspect and analyze inversions from alternative architechtures such as Vision Transformers \cite{dosovitskiy2020image} and MLP-Mixers \cite{tolstikhin2021mlp}. 


We hope that these questions inspire further work which improves learning systems. 

\section{Acknowledgements}

The authors thank Alexey Dosovitskiy for helpful feedback on an earlier draft of this manuscript. The authors also thank the Boston University IVC group for feedback prior to submission. Piotr Teterwak was partially supported by the DARPA XAI program. 

\input{appendix}

\newpage
\bibliographystyle{apalike}
\bibliography{references}

\end{document}

%% file: appendix.tex

%
%
\title{Supplementary Material\\[0.2em]\normalsize Understanding Invariance via Feedforward Inversion\\ of Discriminatively Trained Classifiers}
\date{}
\vspace{-30pt}

\maketitle

\vspace{-30pt}

\section{Noise resampling}

\begin{figure}[h]
  \centering
  \includegraphics[width=1\linewidth]{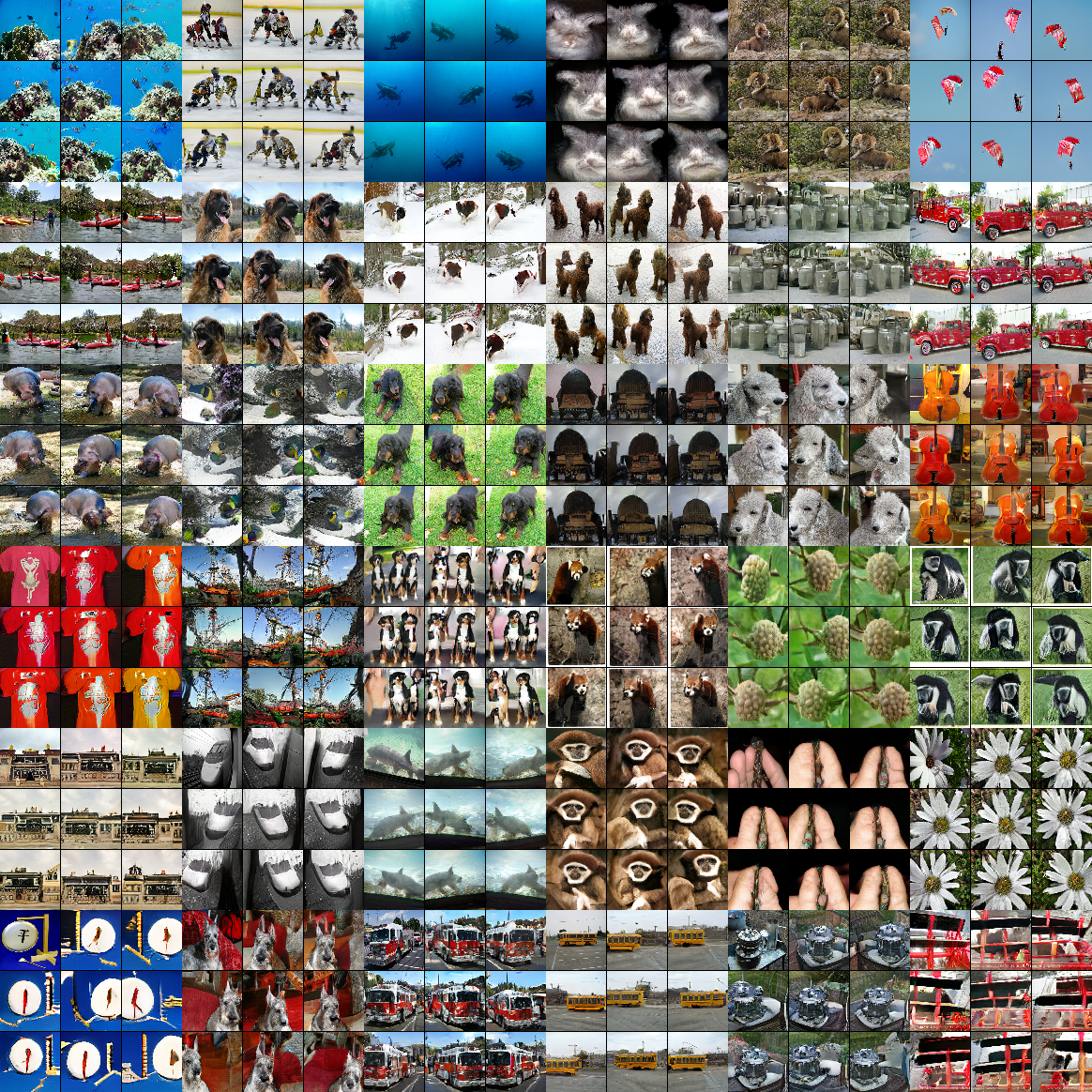}
  \caption{Additional noise resamplings for the robust ResNet-152 on randomly chosen images from the ImageNet validation set.}
  \label{fig:robust_resampling}
\end{figure}

\begin{figure}[h]
  \centering
  \includegraphics[width=1\linewidth]{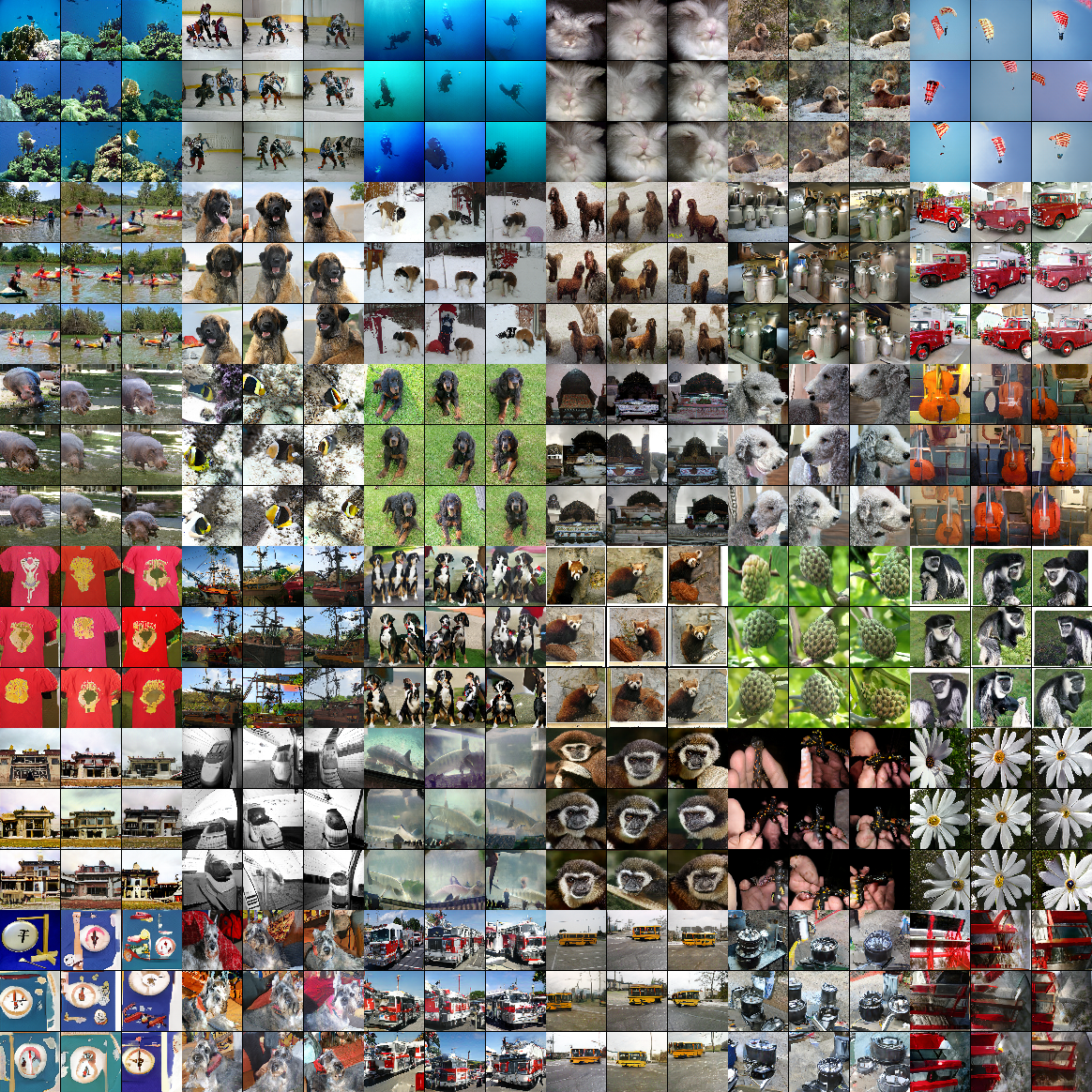}
  \caption{Additional noise resmaplings for the non-robust ResNet-152. }
  \label{fig:non_robust_resampling}
\end{figure}

In Figures \ref{fig:robust_resampling} and \ref{fig:non_robust_resampling}, we provide additional noise resampling results for both robust and non-robust models. For each block of 9 images, the upper left is the input to the classifier. The next 8 are the same logits with different noise vectors fed into the generator. 

\section{Interpolations between logits }

\begin{figure}[h]
  \centering
  \includegraphics[width=1\linewidth]{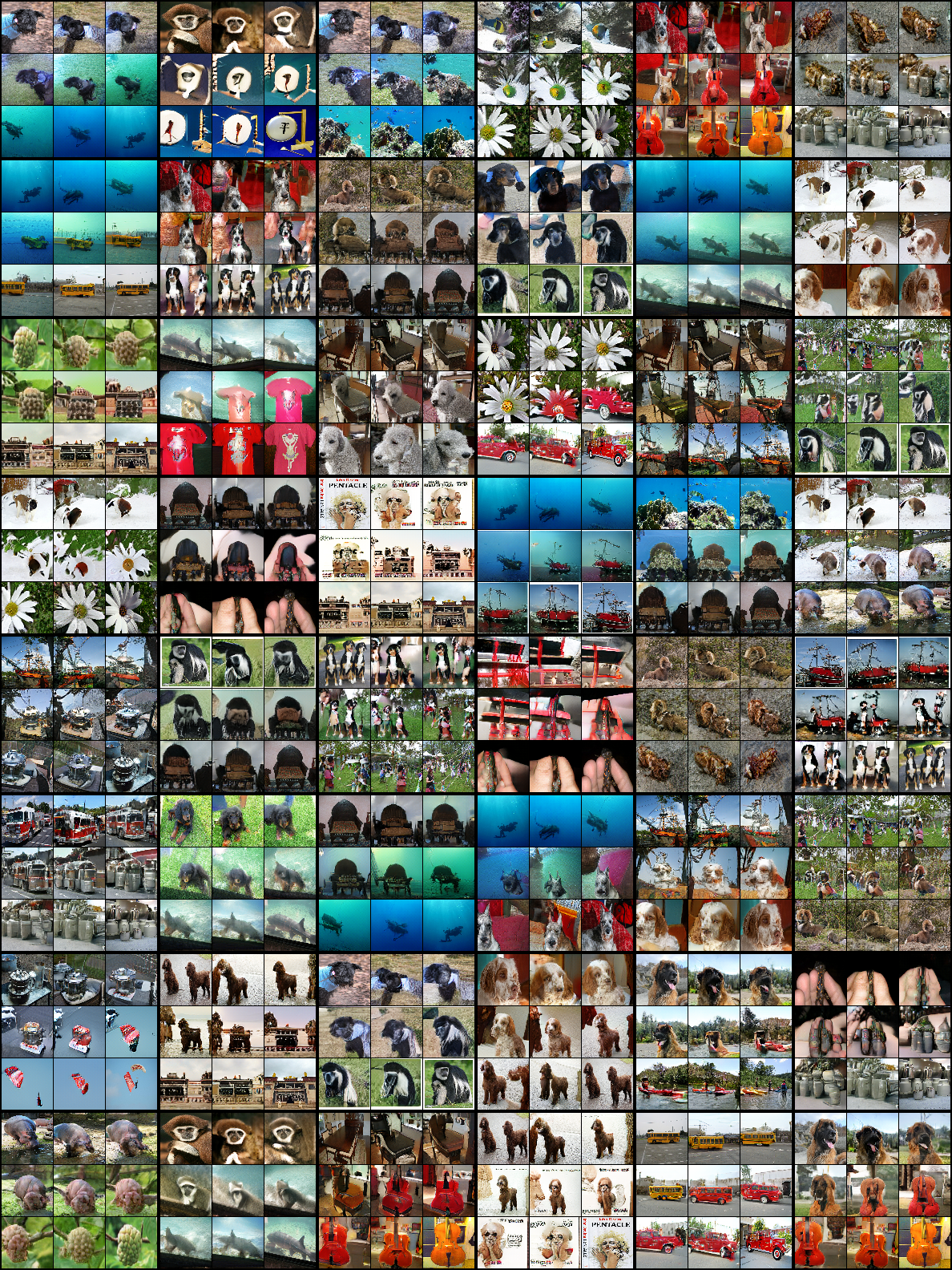}
  \caption{ Linear interpolations between two logit vectors for the robust ResNet-152. Upper-left and lower-right are ground truth images.  }
  \label{fig:robust_interpolations}
\end{figure}

\begin{figure}[h]
  \centering
  \includegraphics[width=1\linewidth]{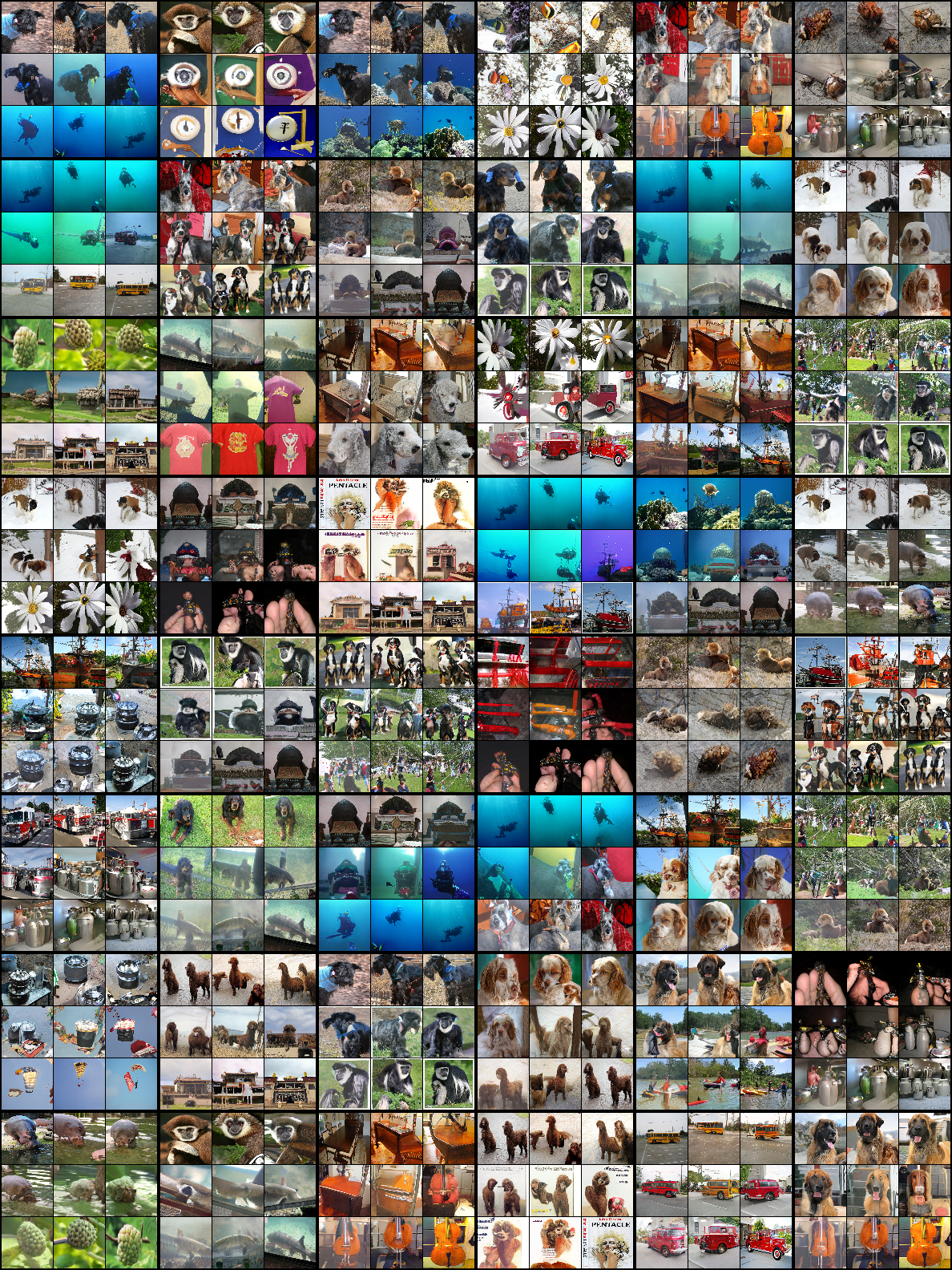}
  \caption{ Linear interpolations between two logit vectors for the non-robust ResNet-152. Upper-left and lower-right are ground truth images.}
  \label{fig:non_robust_interpolations}
\end{figure}

In Figures \ref{fig:robust_interpolations} and \ref{fig:non_robust_interpolations} and we interpolate between two logits of two vectors linearly.  We find robust models to give a smoother sequence. 

\section{Interpolations between noise vectors }

\begin{figure}[h]
  \centering
  \includegraphics[width=1\linewidth]{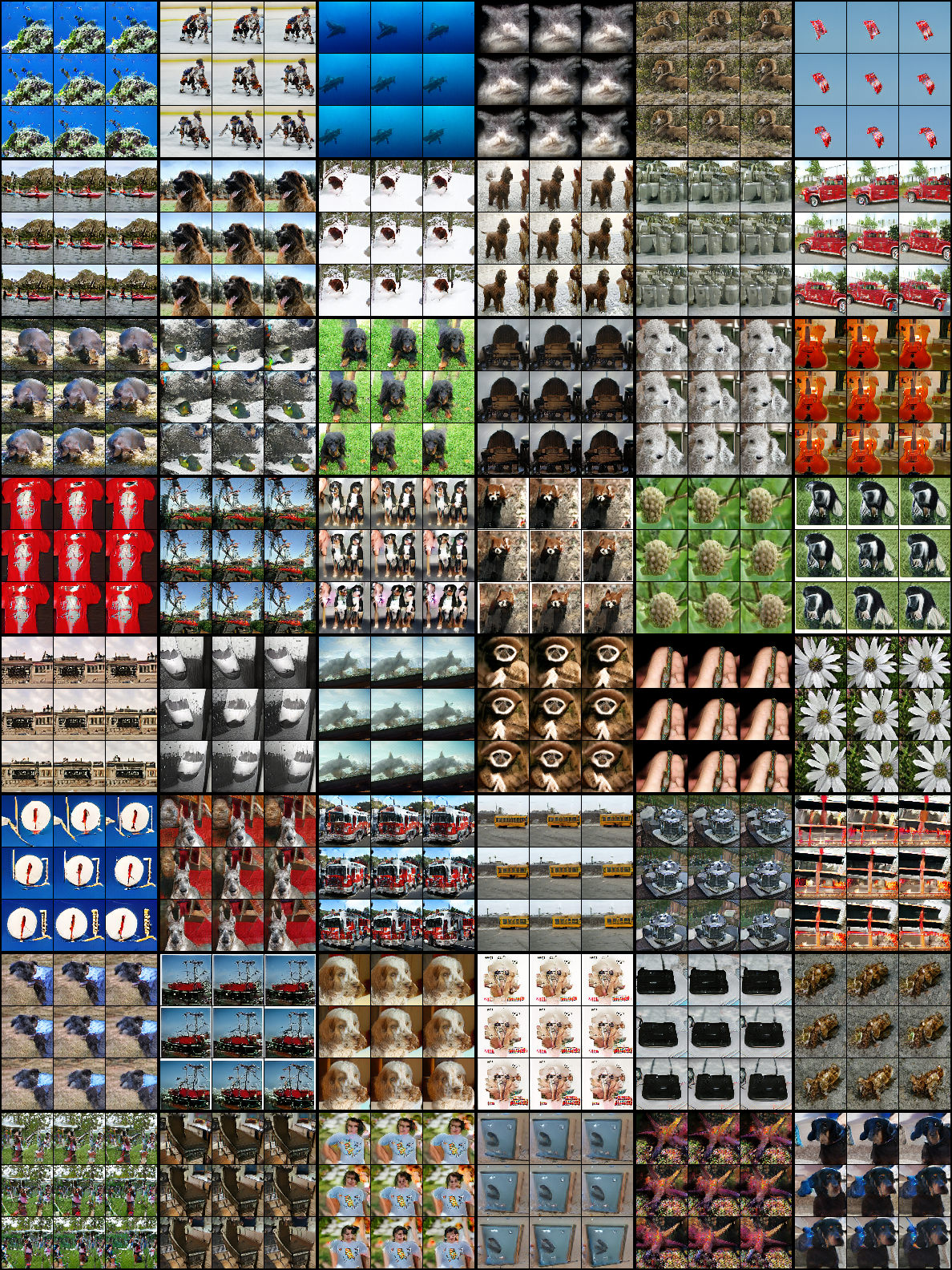}
  \caption{Linear interpolations between two noise inputs for the robust ResNet-152}
  \label{fig:robust_noise_interpolations}
\end{figure}

\begin{figure}[h]
  \centering
  \includegraphics[width=1\linewidth]{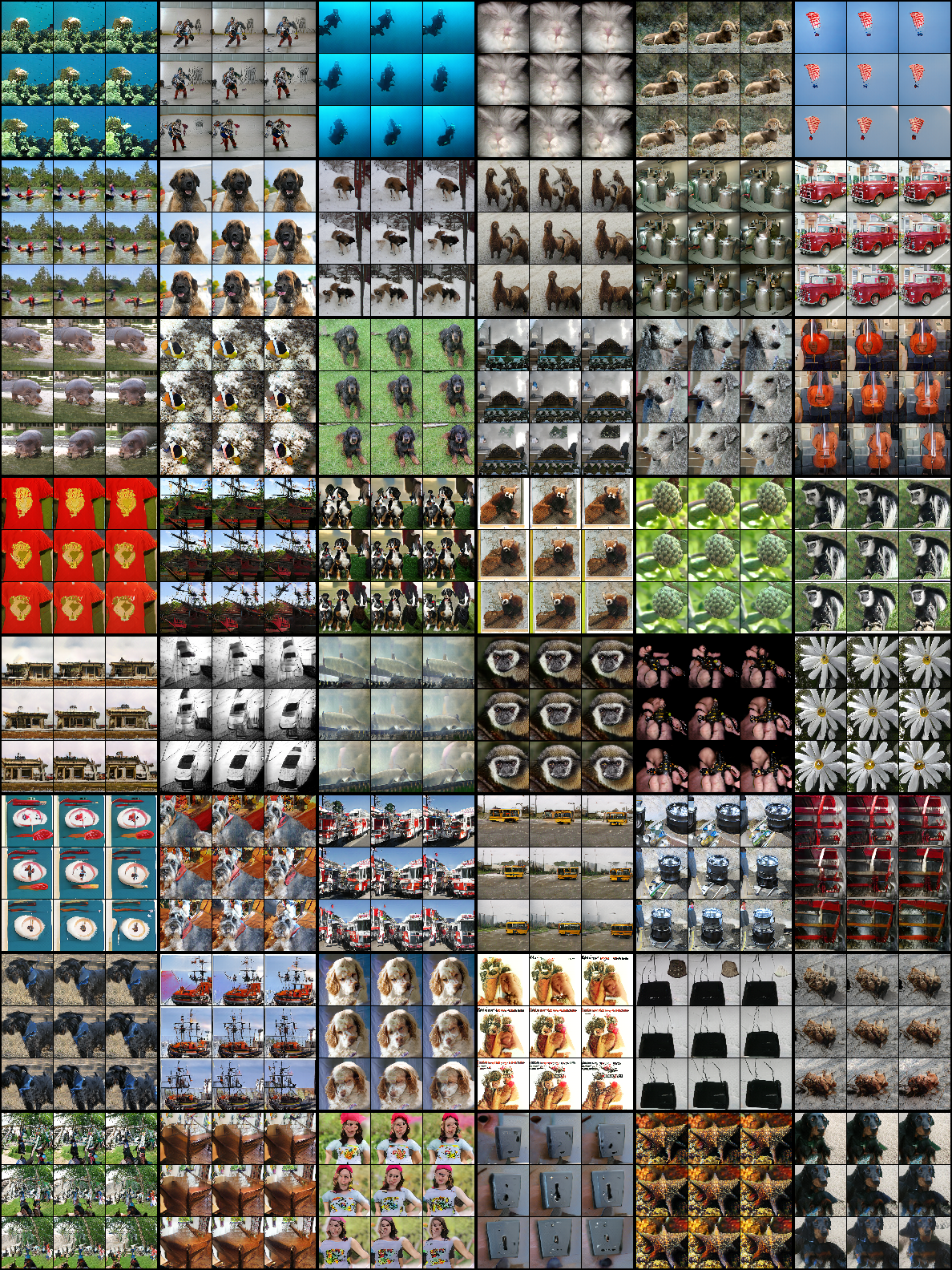}
  \caption{Linear interpolations between two noise inputs for the non-robust ResNet-152}
  \label{fig:non_robust_noise_interpolations}
\end{figure}

In Figures \ref{fig:robust_noise_interpolations} and \ref{fig:non_robust_noise_interpolations} and we interpolate between two  noise vectors linearly. We find that the robust model has a smaller semantic difference between the two endpoints.  

\section{Reconstructing Incorrectly Classified Images}

\begin{figure}[h]
  \centering
  \includegraphics[width=1\linewidth]{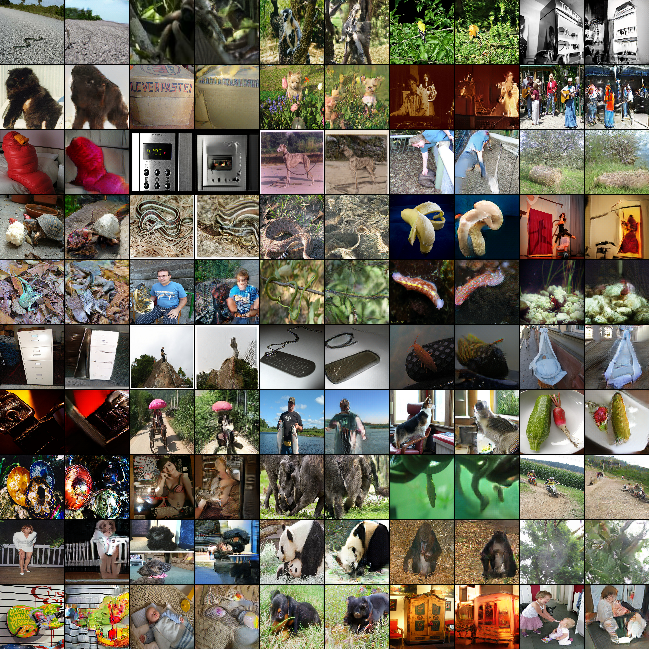}
  \caption{ Reconstructions of incorrectly classified images from the robust ResNet-152.}
  \label{fig:correct_incorrect_robust}
\end{figure}

\begin{figure}[h]
  \centering
  \includegraphics[width=1\linewidth]{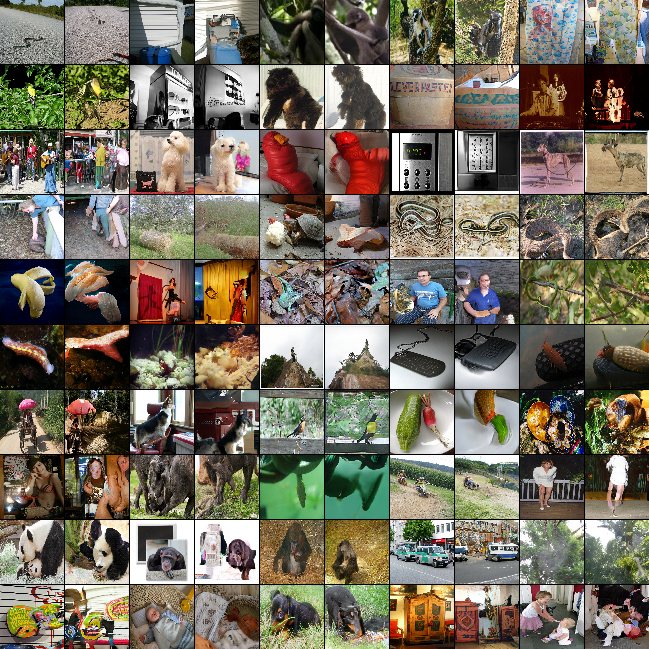}
  \caption{ Reconstructions of incorrectly classified images from the non-robust ResNet-152.}
  \label{fig:correct_incorrect_non_robust}
\end{figure}

In figures \ref{fig:correct_incorrect_robust} and \ref{fig:correct_incorrect_non_robust} we show additional examples for incorrectly classified samples. We find that even for incorrectly classified samples, reconstructions are of high quality. 

\section{Iterative Stability}

We explore iterative stability of the model; if we pass the reconstructed sample back through the classifier, how often does it predict the same class as the ground truth image? The results are much different for the robust and non-robust models. We take 500 samples from 50 classes of ImageNet for this experiment. For the robust model; if the ground truth sample was correctly classified, the reconstruction is classified identically as the ground truth 54\% of the time. If the ground truth was incorrectly classified, the reconstruction is classified identically as the ground truth 35\% of the time. There is a substantial difference between correctly classified samples, despite qualitative observations that reconstructions are similarly good for both. The results for the non-robust model are 49.3\% and 29.9\% respectively. The robust model is more stable, implying that the inversion model is better at preserving features used for classification with the robust model.  

\section{Analysis of scale and shift effects}

\begin{figure}[t!]
  \centering
  \includegraphics[width=0.5\linewidth]{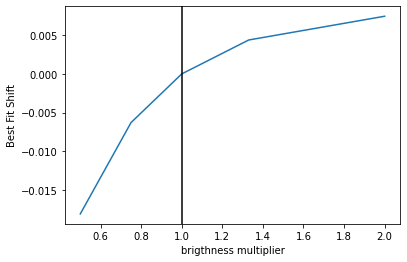}
  \caption{ Plotting best-fit shifts vs. brightness factor for Robust ResNet-152. As brightness increases, so does expected shift. Brightness factor of 1 corresponds to the identity function, so expected shift is 0, this is represented by the vertical line.}
  \label{fig:best_fit_shift}
\end{figure}

\begin{figure}[h]
  \centering
  \includegraphics[width=0.5\linewidth]{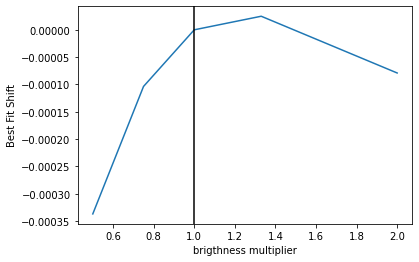}
  \caption{ Plotting best-fit shifts vs. brightness factor, for non-robust ResNet-152. There is a much weaker relationship than for the robust ResNet-152 model. Brightness factor of 1 corresponds to the identity function, so expected shift is 0, this is represented by the vertical line.}
  \label{fig:best_fit_shift_non_robust}
\end{figure}

In the main paper, we show that manipulating logits before they are input to inversion model resulted in systematic changes; both scale and shift seemed to encode brightness for the robust ResNet-152 model, while it didn't have as pronounced an effect on the non-robust model. Here, we show exactly this experimentally for shifts. We start with a batch of images, and then multiply its pixel values by a fixed value to adjust the brightness. Then for each brightness-adjusted image, we compute robust logits. We then find a best-fit shift to get back to the non-brightness adjsuted logits. In other words, we fit the shift parameter in the mode.: $logits_{original} = logits_{brightness-adjusted} + shift$. We do this independently for each image, and report the mean across brightness factors in Figure \ref{fig:best_fit_shift} We find that as the brightness increases , the best-shift does as well. This verifies the finding in the main paper. However, for the non-robust model, the expected shift peaks and then decreases, shown in \ref{fig:best_fit_shift_non_robust}. 

\begin{figure}[h]
  \centering
  \includegraphics[width=0.5\linewidth]{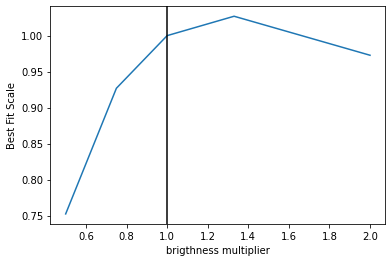}
  \caption{ Plotting best-fit scales vs. brightness factor, for robust ResNet-152. Brightness factor of 1 corresponds to the identity function, this is represented by the vertical line. Locally, around this line, there is a positive relationship between scale and brightness.}
  \label{fig:best_fit_scale}
\end{figure}

\begin{figure}[h]
  \centering
  \includegraphics[width=0.5\linewidth]{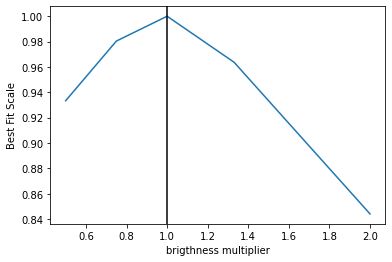}
  \caption{ Plotting best-fit scales vs. brightness factor, for non-robust ResNet-152. Brightness factor of 1 corresponds to the identity function, this is represented by the vertical line.Locally, around this line, there is no positive relationship between scale and brightness.}
  \label{fig:best_fit_scale_non_Robust}
\end{figure}

For scale, we perform a similar experiment, except we fit the scale in the model $logits_{original} = logits_{brightness-adjusted} * scale$. We find that the relationship for scale is weaker than for shift. However, locally around the identity transform, there is still a positive relationship between scale and brightness for the robust ResNet-152 (Figure \ref{fig:best_fit_scale}); the best-fit scale peaks at a greater brightness than the identity transform. On the other hand, for the non-robust ResNet-152 (Figure \ref{fig:best_fit_scale_non_Robust}), the identity transform represents the peak expected scale of 1. 

\begin{figure}[h]
  \centering
  \includegraphics[width=0.5\linewidth]{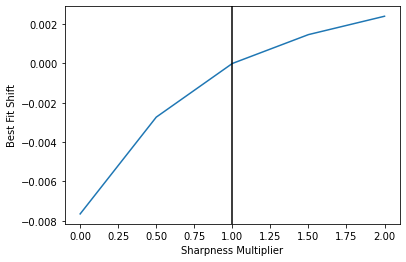}
  \caption{ Plotting best-fit shifts vs. sharpness factor, for Robust ResNet-152. Sharpness factor of 1 corresponds to the identity function, so expected shift is 0, this is represented by the vertical line.}
  \label{fig:best_fit_shift_sharpness}
\end{figure}

\begin{figure}[h]
  \centering
  \includegraphics[width=0.5\linewidth]{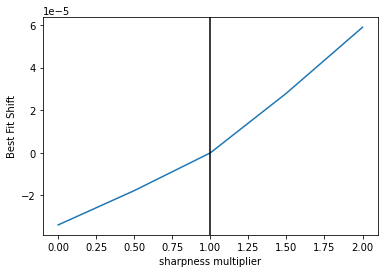}
  \caption{ Plotting best-fit shifts vs. sharpness factor, for non-robust ResNet-152.  Sharpness factor of 1 corresponds to the identity function, so expected shift is 0, this is represented by the vertical line. There is a positive relationship.}
  \label{fig:best_fit_shift_non_robust_sharpness}
\end{figure}

\begin{figure}[h]
  \centering
  \includegraphics[width=0.5\linewidth]{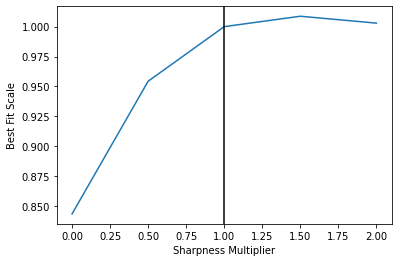}
  \caption{ Plotting best-fit scales vs. sharpness factor, for robust ResNet-152. Brightness factor of 1 corresponds to the identity function, this is represented by the vertical line. Although there is a positive relationship, the best fit shifts are much smaller magnitude. }
  \label{fig:best_fit_scale_sharpness}
\end{figure}

\begin{figure}[h]
  \centering
  \includegraphics[width=0.5\linewidth]{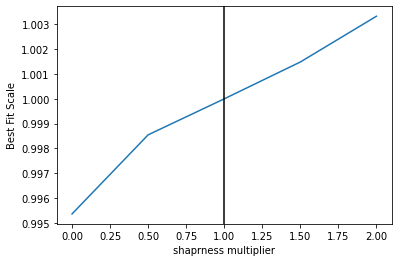}
  \caption{ Plotting best-fit scales vs. sharpness factor, for non-robust ResNet-152. Sharpness factor of 1 corresponds to the identity function, this is represented by the vertical line. There is a strong relationship. Again, although there is a positive relationship, the expected scales magnitudes are much smaller than for the robust model.}
  \label{fig:best_fit_scale_non_Robust_sharpness}
\end{figure}

We repeat the experiment for image sharpness manipulations. Image sharpness of 0 means a smoothed image, while image sharpness of 2 means a sharpened image. We show results for the Robust ResNet-152 in Figures \ref{fig:best_fit_shift_sharpness} and \ref{fig:best_fit_scale_sharpness}, and non-robust ResNet-152 in Figures \ref{fig:best_fit_shift_non_robust_sharpness} and \ref{fig:best_fit_scale_non_Robust_sharpness}. 

We find that in both models, best-fit scale and shift positively correlate with sharpness. However, for the Robust model, the magnitudes of expected scales and shifts is much much larger.

\subsection{Effect of image-space rotations on reconstructions}

\begin{figure}[bt!]
  \centering
  \includegraphics[width=1.0\linewidth]{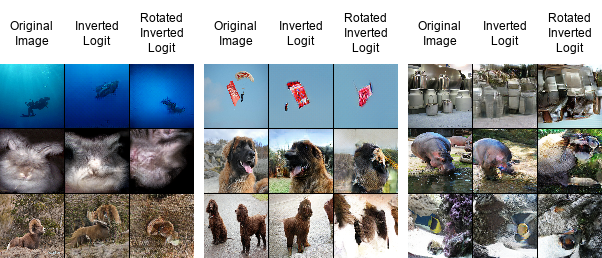}
  \caption{Reconstruction of rotations. We compare images encoded with Robust ResNet-152 (left column) to the their reconstructions (middle), and reconstructions of the same image rotated by 90 degrees (right). We counter-rotate the reconstruction for ease of comparison. The rotated samples reconstruct textures reasonably well but objects are horribly disfigured.}
  \label{fig:rotations}
\end{figure}

In the main paper, we explored manipulations of the logits, i.e., the input to the decoder. 
With logit manipulations, we can infer what aspects of the image will cause the classifier to produce corresponding variation in the logits.
In this section, we directly manipulate images, in order to determine whether the robust classifier preserves information about the manipulation and whether
the decoder can recover the original image. The particular manipulation we study is rotation of the image by 90 degrees.  Figure~\ref{fig:rotations}
shows an original image, the reconstruction of this image from the robust logits, and the reconstruction of this image rotated by 90 degrees. To make it easier
to compare images, we  counter-rotate the latter reconstruction so that it will be identical to the original image if it is reconstructed veridically.
We observe some signal in the rotated reconstruction, but clearly there is degradation. For example, animals---which are nearly always upright in 
ImageNet---are horribly distorted. The classifier and/or decoder do not appear to generalize well outside of the data distribution they were trained 
on (i.e., upright data). Interestingly, textures are fairly well reconstructed, such as the background brush in the third row, left triplet, which 
makes sense given that textures tend not to have a reliable orientation in the natural world.

\section{Training Details}

\subsection{Network Architectures}

We use a BigGAN generator configured for 64x64 images. The layer configuration is listed below. Each residual block contains 3 conditional-batch norms and a single 2x upsampling operation, in addition to convolutions and non-linearities.  We use hierarchical noise input, detailed in \cite{brock2018large}. For more details, please see \cite{brock2018large} and the \href{https://github.com/google/compare_gan}{compare\_gan} codebase. 

\begin{table}[h!]
\small
\setlength{\tabcolsep}{4pt}
\centering
\begin{tabular}{|c|c|c|}
\hline
Layer & Out Resolution  & Out Channels  \\
\hline \hline
Linear(noise) & 4 & 1536 \\
\hline
Residual Block Upsampling & 8 & 1536 \\
\hline
Residual Block Upsampling & 16 & 768 \\
\hline
Residual Block Upsampling & 32 & 384 \\
\hline
Residual Block Upsampling & 64 & 192 \\
\hline
Non-local block (self-attention) & 64 & 192\\
\hline
BN, ReLU, and Conv & 64 & 3 \\
\hline
Tanh & 64 & 3 \\
\hline

\end{tabular}
\caption{Generator architecture}
\end{table}

Similarly, we use the BigGAN discriminator. There are 4 resblock with downsampling, and one without. The proection discriminator, described in the main paper, comes at the end. Again, for  more details, please see \cite{brock2018large} and the \href{https://github.com/google/compare_gan}{compare\_gan} codebase.

\begin{table}[h!]
\small
\setlength{\tabcolsep}{4pt}
\centering
\begin{tabular}{|c|c|c|}
\hline
Layer & Out Resolution  & Out Channels  \\
\hline \hline
ResBlock Down & 32 & 192 \\
\hline
Non-local block (self-attention) & 32 & 192 \\
\hline
ResBlock Down & 16 & 384 \\
\hline
ResBlock Down & 8 & 768 \\
\hline
ResBlock Down & 4 & 1536 \\
\hline 
ResBlock & 4 & 1536 \\
\hline
ReLU, Global Sum pooling & 1 & 1536\\
\hline
Conditional Projection & 1 & 1 \\
\hline
\end{tabular}
\caption{Discriminator architecture}
\end{table}

We train with the Adam optimizer with $\beta_1$ set to 0 and $\beta_2$ set to 0.999, and a learning rate of $0.0001$ for the generator and $0.0005$ for the discriminator. We take 2 discriminator steps for every generator step. We initialize weights orthognally, and use spectral norm in both the generator and discriminator. For more details, see below. 

\subsection{Training Configuration}
\label{appendix:config}

Below, the exact configuration used for the inversion models in the paper. It can be used directly with the \href{https://github.com/google/compare_gan}{compare\_gan} code.

{\tiny
\begin{verbatim}
# Parameters for AdamOptimizer: 
# ===============================================
AdamOptimizer.beta1 = 0.0 
AdamOptimizer.beta2 = 0.999 

# Parameters for D: 
# ===============================================
D.spectral_norm = True 

# Parameters for dataset: 
# ===============================================
dataset.name = 'imagenet_64' 

# Parameters for resnet_biggan.Discriminator: 
# ===============================================
resnet_biggan.Discriminator.project_labels = True 

# Parameters for eval_z: 
# ===============================================
eval_z.distribution_fn = @tf.random.stateless_normal 

# Parameters for G: 
# ===============================================
G.batch_norm_fn = @conditional_batch_norm 
G.spectral_norm = True 

# Parameters for resnet_biggan.Generator: 
# ===============================================
resnet_biggan.Generator.embed_labels = True 
resnet_biggan.Generator.hierarchical_z = True 

# Parameters for loss: 
# ===============================================
loss.fn = @hinge 

# Parameters for ModularGAN: 
# ===============================================
ModularGAN.conditional = True 
ModularGAN.d_inputs = ['label'] 
ModularGAN.d_lr = 0.0005 
ModularGAN.d_optimizer_fn = @tf.train.AdamOptimizer 
ModularGAN.g_inputs = ['label'] 
ModularGAN.g_lr = 0.0001 
ModularGAN.g_optimizer_fn = @tf.train.AdamOptimizer 
ModularGAN.g_use_ema = True 

# Parameters for options: 
# =================================================
options.architecture = 'resnet_biggan_arch' 
options.batch_size = 2048 
options.disc_iters = 2 
options.gan_class = @ModularGAN 
options.lamba = 1 
options.training_steps = 250000 
options.z_dim = 120 

# Parameters for penalty: 
# =================================================
penalty.fn = @no_penalty 

# Parameters for run_config: 
# =================================================
run_config.iterations_per_loop = 500 
run_config.save_checkpoints_steps = 2500 
run_config.tf_random_seed = 8 

# Parameters for spectral_norm: 
# =================================================
spectral_norm.singular_value = 'auto' 

# Parameters for standardize_batch: 
# =================================================
standardize_batch.decay = 0.9 
standardize_batch.epsilon = 1e-05 
standardize_batch.use_moving_averages = False 

# Parameters for weights: 
# =================================================
weights.initializer = 'orthogonal' 

# Parameters for z: 
# =================================================
z.distribution_fn = @tf.random.normal 
\end{verbatim}}

